# State-of-the-Art Transformer Models for Image Super-Resolution: Techniques, Challenges, and Applications

**Debasish Dutta**[1*], Deepjyoti Chetia [2], Neeharika Sonowal [3] and Sanjib Kr Kalita [4]

[1,2,3,4]Dept of Computer Science, Gauhati University, Assam, India

*Corresponding Author: **Debasish Dutta**          Email: debasish@gauhati.ac.in.

**Abstract:**

Image Super-Resolution (SR) aims to recover a high-resolution image from its low-resolution counterpart, which has been affected by a specific degradation process. This is achieved by enhancing detail and visual quality. Recent advancements in transformer-based methods have remolded image super-resolution by enabling high-quality reconstructions surpassing previous deep-learning approaches like CNN and GAN-based. This effectively addresses the limitations of previous methods, such as limited receptive fields, poor global context capture, and challenges in high-frequency detail recovery. Additionally, the paper reviews recent trends and advancements in transformer-based SR models, exploring various innovative techniques and architectures that combine transformers with traditional networks to balance global and local contexts. These neoteric methods are critically analyzed, revealing promising yet unexplored gaps and potential directions for future research. Several visualizations of models and techniques are included to foster a holistic understanding of recent trends. This work seeks to offer a structured roadmap for researchers at the forefront of deep learning, specifically exploring the impact of transformers on super-resolution techniques.

## 1. Introduction

Super-resolution (SR) is the process of amplifying Low-Resolution (LR) images to High-Resolution (HR). The applications range from natural images to medical imaging to compressed images and enhancement to highly advanced satellite and medical imaging.

### 1.1. Background

There can be many types of SR, like generating SR from a single image (SISR) or multiple images (MISR). Also, some SR models are trained to take a reference image along with the LR input (RefSR) to obtain the final HR image. [3]

Despite notable achievements of prior SR models, SR remains a challenging task in computer vision because it is notoriously ill-posed like several HR images can be valid for any given LR image due to many aspects like brightness and coloring. Traditionally, SR was performed using mathematical means, and after the advent of Deep Learning, DL-based methods [4] like CNNs and GANs took over. Then, throughout numerous advancements, attention was introduced, which led to the development of the Transformer [29] and thus began the rapid advances in the field of SR image generation, thereby solving limitations of previous methods like limited receptive fields, poor global context capture, and difficulty in high-frequency detail recovery. This work unwinds some of the most recent advances in the field of SR.

### 1.2. Related Works

Although many surveys have been conducted in the field of SR, most of them focused on the conventional algorithms as [1], [2]. [1] discussed the bases of almost all of the previously existing SR algorithms and also proposed a detailed taxonomy of the algorithms, and divided them into spatial and transform domains as well as single-image and multiple-image algorithms. Wang et al. [2] specifically focused on Single image super-resolution (SISR), and they evaluated the State-of-the-art SISR methods using two benchmark datasets of that time. After the rise of deep learning, traditional SR models have mostly been overtaken by DL-based models. Yang et al. [3] provided an overall review of SR models using DL, focusing on efficient architecture designs and well-defined optimization objectives. Meanwhile, Wang et al. [4] conducted a comprehensive survey on DL models, offering a structured classification of existing models categorized into Supervised, Unsupervised, and domain-specific applications. More recently, Moser et al. presented two surveys on SR [6, 7], discussing different learning strategies, mechanisms, and architectures used by diverse SR models. Their follow-up paper included information about diffusion models and their integration into SR models. Although many existing surveys exist, there has not been any consolidated work that discusses the Transformer network's adaptation to the task of SR and the groundbreaking transformer-based [8] SR models. This study thereby seeks to fill this gap by providing an in-depth overview of the adaptation of transformers for generating super-resolved images and a review of the state-of-the-art SR methods using transformer networks. This work also aims to discuss potential applications and highlight challenges along with its future directions.

### 1.2. Contribution of the work

A. The primary contribution of this work is to provide insight into the recent developments in the field of super-resolution imaging using SOTA transformer networks. B. The study also discusses gaps with potential improvements in the field.

### 2. SR Problem Definition and Setting

### 2.1. Problem Definition of SR Task

For an LR image $x \in \mathbf{R}^{h \times w \times c}$, the goal of an SR model is to generate the associated HR image $y \in \mathbf{R}^{\bar{h} \times \bar{w} \times c}$, which $h, w$ represents the height and width of the image with $c$ channels $h < \bar{h}$ and $w < \bar{w}$. This can be mapped as:

$$x = \mathbf{D}(\bar{x}; \Theta) = \left((y \otimes k)_{\downarrow_s} + n\right)_q \dots\dots\dots\dots\dots\dots\dots\dots\dots\dots\dots\dots (1)$$

Where, $\mathbf{D}$ is the degradation map $\mathbf{D}: \bar{x} \in \mathbf{R}^{\bar{h} \times \bar{w} \times c} \to \bar{x} \in \mathbf{R}^{h \times w \times c}$ and $\Theta$ represents the set of degrading parameters blur kernel $k$, noise $n$, scaling factor $s$ and compression quality $q$.[9] Since the degradation is mostly unknown, this formulates the primary challenge of determining the inverse $D$ along with its parameters $\theta$. Thus, the primary objective of an SR model $\mathbf{M}: \bar{x} \in \mathbf{R}^{\bar{h} \times \bar{w} \times c} \to y \in \mathbf{R}^{h \times w \times c}$ is to inverse Eq1 $\hat{y} = \mathbf{M}(x; \theta)$, where $\hat{y}$, which is the HR approximation of the provided LR image $x$ with $\theta$ degradation parameters. For a DL model, this thereby becomes an optimization problem which minimizes the difference between the estimated HR image $\hat{y}$ And the ground truth $y$ and a loss function $\mathfrak{L}$:

$$\hat{\theta} = argmin_\theta \, \mathfrak{L}(\hat{y}, y) + \lambda_\phi(\theta) \dots\dots\dots\dots\dots\dots\dots\dots\dots\dots\dots\dots (2)$$

where, $(\phi(\theta)$ is the regularization term weighted by $\lambda$.

### 2.2. Learning in Super-Resolution

In an SR task, the learning functions differ from that of a traditional model, which is for high-level tasks such as detection and classification. The sub-section briefly discusses some of the common loss functions.

#### 2.2.1. Pixel Loss

Pixel loss [10] is one of the major loss functions used while training an SR network. It is calculated as the difference between the pixels of the ground truth reference image and the reconstructed super-resolved image. Generally, Mean Square Error (MSE) also called $L_1$ loss is used as the difference, but some have found better results using Mean Absolute Error (MAE), also termed $L_2$. In super-resolution, training with pixel loss increases the PSNR but does not have any direct correlation to the perceived image quality.

### 2.2.2. Perceptual Loss

Perceptual loss [11] tends to capture the high-level features in the generated HR image with that of the provided ground truth, which the pixel-based loss function often lacks. This $\mathcal{L}_{per}$ is done by calculating the difference in feature maps of the two images. Perceptual loss helps the generator capture semantic and structural similarities rather than focusing solely on pixel-level differences, which leads to more visually accurate results.

### 2.2.3. Adversarial Loss

For GAN-based methods, a different loss function is introduced called adversarial loss [12], which is based on the mechanics between the generator (G) and discriminator (D) network. This $\mathcal{L}_{adv}$ penalizes D for the input gradient, which helps stabilize the training of a GAN, generating high-quality images with faster convergence.

### 2.2.4. Texture Loss

Texture loss similar to perceptual loss, is designed to preserve the fine-grained texture details, which are often missed by the $\mathcal{L}_{per}$ by comparing texture patterns between the generated and ground-truth images. It is often defined using Gram matrices, as inspired by style transfer [13].

### 2.2.5. Combined Loss for Super-Resolution

In practice, SR models often combine all of these losses into a single loss function.

$$\mathcal{L} = \mathcal{L}_{pixel} + \lambda_{adv}\mathcal{L}_{adv} + \lambda_{tex}\mathcal{L}_{tex} + \lambda_{per}\mathcal{L}_{per} \quad\quad\quad\quad\quad (3)$$

Where $\mathcal{L}_{pixel}$ is typically an $L1$ or $L2$ loss and $\lambda_{adv}, \lambda_{tex}, \lambda_{perc}$ are the regularization weights. This combination ensures both pixel similarity and high-level perceptual quality.

### 2.3. Evaluation: Image Quality Assessment

Evaluating an SR reconstructed image requires specialized methods to determine how realistic the image appears after applying SR methods. Quantitative methods like PNSR and SSIM [14] use a mathematical foundation to calculate the pixel difference between the LR and HR images. Along with the quantitative analysis, qualitative analysis is also critical in a study. They create a more robust understanding of the results, thus enabling better research analysis like Mean Opinion Square, which is a subjective analysis metric where human subjects rate the visual quality of images based on their perceptual opinion. [16] Learning Perceptual Quality [17], on the other hand, uses ML models to evaluate the perceptual quality of images.

### 2.4. Datasets & Challenges

A diverse range of image datasets are available to be used for SR, each offering unique qualities. These datasets vary in resolution, image count, and content, with some providing high-resolution images ideal for detailed SR tasks but requiring significant computational resources. The most commonly used dataset for training SISR algorithms is DIV2k [30], and an extended version of it is called DF2K. And CUFED5 [31] for RefSR. For testing, there exist multiple benchmarking datasets having a wide collection of images across multiple domains. The most famous benchmarking challenges are New Trends in Image Restoration (NITRE), part of the CVPR[5], and there are also workshops by CVF where most of the SOTA algorithms are introduced.

### 3. Review of SOTA Transformers for SR

In this section, a brief overview of some state-of-the-art transformer-based methods is discussed. A comparison among SOTA models is given in Table 1 and one on their number of parameters used vs PSNR is shown the Figure 1.

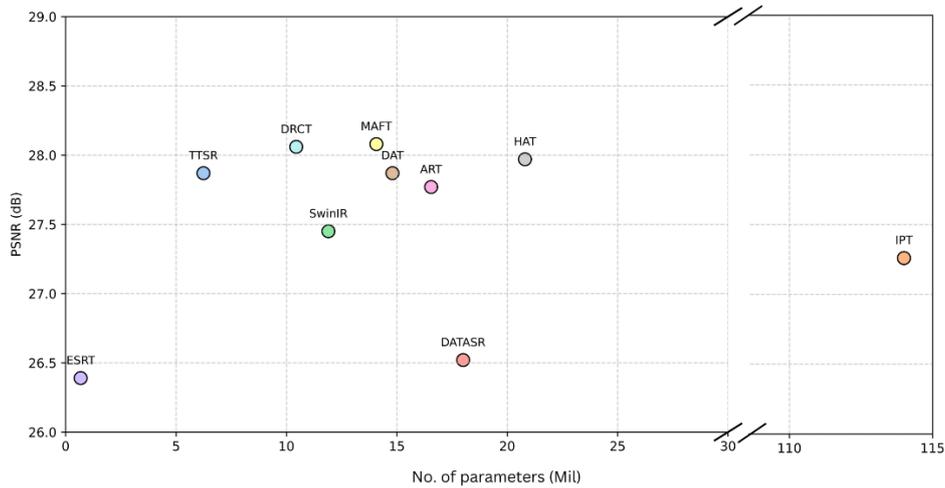

Figure 1: A comparison between the SOTA methods of their no. of parameters used vs PSNR

### 3.1. Early Pioneering Works

After realizing the tremendous success of Transformers [15] in NLP and Vision, researchers have started to adopt the base methodology into low-vision works like image SR, denoising, and such. In the further sections, the breakthrough methods are discussed briefly.

Even before the introduction of ViT, Yang et al. [18] proposed this texture transformer architecture based on the traditional transformer to address the problem of ineffective learning of high-level semantic features. Being a RefSR model, along with an associated LR image, their proposed Texture Transformer also takes a Ref image as input to output a synthesized feature map that is further used to generate the predicted HR image. Additionally, they also proposed a cross-scale feature integration module that stacks these texture-transformers, achieving better feature representation across different SR scales (1 × to 4 ×), and this proved to be quite useful in passing relevant features across different tasks.

One of the pioneer works in using the transformer architecture for low-vision tasks is this novel architecture by Chen et al. [19]. They developed a pre-trained model architecture formed up of four blocks - heads that extract the features from the degraded input images, an encoder-decoder transformer used for restoring the missing information with the data, and finally, tails that map the features onto the restored image. They used a multi-head and multi-tail to deal with the task separately. The encoder-decoder part here, instead of using the traditional approach, uses the method depicted by ViT [29]. They split the input features into a series of patches. The base architecture of the encoder layer follows the same structure as the ViT, consisting of a multi-head self-attention (MSA) module and a feed-forward network (FFN). The decoder also follows the same architecture and hence takes the output in the transformer body, which has two MSA layers and one FFN. The only modification they did was the addition of an auxiliary input to the decoder, a task-specific embedding that learns the decoding feature for each task.

Liang et al. [20] introduced the image restoration model SwinIR, based on the Swin Transformer, following the promising results of Swin [32] on high-level tasks. Since Swin combines both CNN and Transformer, it captures large image feature size due to local attention and captures long-range dependencies through the Transformer. SwinIR has three basic modules: shallow feature extraction, deep feature extraction, and HR reconstruction. Many others later employed this structure. The shallow feature extraction module uses a convolutional layer to extract shallow features, which are directly passed to the reconstruction module. The deep feature extraction module comprises residual Swin Transformer blocks

(RSTB), with each block utilizing multiple Swin Transformer Layers (STL) for local attention and cross-window interaction. This use of STLs and a convolutional layer enhances the translational equivariance of SwinIR, which generic transformers lack due to their nature as specific instantiations. Additionally, the residual connection helps create a short identity-based link between the reconstruction module, allowing aggregation at different feature levels. The STL is based on the original Transformer layer but has a primary distinction in local attention and the shifting window mechanism. This shifting window enables the computation of general self-attention in parallel, which is then concatenated for multi-head self-attention (MSA). Importantly, with a predetermined partition size, no connections occur across local windows among different layers. Therefore, regular and shifted window partitioning is used, which alternatively facilitates cross-window connections.

### 3.2. Building on Foundations

After seeing success in adopting the transformer architecture through ViT and Swin, Cao et al. [21] proposed a refSR model with a deformable attention transformer called DATASR, which is built on the U-Net network but with multi-scale features, thereby alleviating the resolution gap between and mollifying the mismatching issues between the LR and Ref images. They used the transformer-encoder to extract the multi-scale features from the ref image. The model matches the correspondences that transfer the textures from ref images to LR ones, also aggregating the features that generate the resultant HR images. This significantly gave better results than that of the TTSR, as DATASR captured the underlying perceptual quality much better compared to its predecessor. This DATASR used a combination of L1, perceptual, and adversarial loss, as given in sec 2.2.

Chen et al., in their work [26], strived to combine two dimensions in a Transformer instead of utilizing self-attention just along one of the dimensions, spatial or channel. DAT proposed to aggregate features across spatial and channel dimensions in an inter-intra-block dual manner. They also follow the method of shallow feature extraction followed by deep feature extraction and, finally, reconstruction of the HR image, which also has a pixel-shuffle branch and a global residual connection (which provides stability). The deep feature extraction consists of Residual Groups (RGs), and each RG contains pairs of dual-aggregation transformer blocks (DATBs). Each DATB pair contains two transformer blocks: dual spatial Transformer block (DSTB) and dual channel Transformer block (DCTB), for the spatial and channel self-attention, respectively. These alternating DSTB and DCTB help DAT collect inter-block feature aggregation across dimensions.

Li et al. analyzed previous transformer-based methods that showed significant results in SR tasks, addressing the inherent problem of long-range dependencies by using local and self-attention mechanisms and cross-layer connections to discover a glaring limitation of the spatial extents of the input feature maps. To solve the issue, the Multi-Attention Fusion Transformer (MAFT) [27] is designed to expand the activated pixel range during image reconstruction, which will effectively utilize more input information space. It improves the balance between the local features and global information, increasing the range and number of activated pixels, leading to a substantial increase in reconstruction performance, along with reducing the reconstruction loss even though there was a substantial increase in the region of pixel utilization in the feature maps.

Most of these above methods increased model performance by expanding the receptive fields or designing deep networks however, Hsu et al. observed that the feature map intensity was suppressive to smaller values towards the tail of the network. They proposed Dense-residual-connected Transformer (DRCT) [28] to mitigate this bottleneck and diminishment of spatial information. DRCT is designed to stabilize the forward propagation and limit feature bottlenecks. For this, they introduced Swin-based Swin-Dense-Residual-connected Block (SDRCB), which encompasses STL and transition layers into Residual DFGs (RDG). This approach enhances the receptive areas with much fewer parameters, thereby improving SISR tasks with more detailed and context-aware processing.

### 3.3. Contemporary Developments

Considering the heavy computational load and exorbitant GPU usage of ViTs, Lu et al. [22] at CVPR22 introduced ESRT, an efficient transformer for SR tasks, which is one of the lightest transformer models to date consuming just over 4GB GPU memory. Even though it has a very low computational cost, it still achieved results comparable to most SOTA SR methods. This hybrid model consists of two important blocks - Lightweight CNN Backbone (LCB) and Lightweight Transformer Backbone (LTB) along with a feature extraction head and image reconstruction tail. The LCB using High Preserving Blocks (HPB) dynamically adjusts the sizes of the feature maps, thus extracting deep features with a very low computational cost. Similarly, LTB captures the long-term dependencies between akin patches of images utilizing specifically designed Efficient Transformer (ET) and Efficient Multi-Head Attention (EMHA). They also validated the effectiveness of ET by implementing ET into RCAN, which reduced the parameters of the original RCAN by almost half to 8.7M from 16M while keeping the performance almost the same or even better in a few cases. This model depicts the best trade-offs between computational cost and model performance.

Chen et al. observed that till then, transformer-based SR networks utilized only a specific spatial range of information, which they aimed to cover up. So, they proposed a hybrid transformer, namely HAT [23], which, to activate more pixel range, combines both channel and self-attention, employing both global as well as local features. They also introduced an overlapping cross-attention module (OCAM), which adds more direct interaction to neighboring feature maps. HAT showed that although the Transformer has a stronger ability to extract local features, its range needed to be expanded. The OCAM computes keys and values over a larger spatial area as compared to SwinIR, which enables better feature aggregation than window-partition-based SwinIR. This larger spatial range of input pixels improves reconstruction accuracy by a huge margin and thus can also be scaled effectively. Till now, HAT has the highest PSNR amongst all the explored SOTA methods.

Zhang et al. argued that where other Sota methods depend on numerous different backbones, using a basic transformer, higher results can still be achieved. Considering the dense attention strategy employed by methods such as SwinIR and IPT using the shifted-window scheme and splitting features into patches, respectively, leads to a restricted receptive field. To address this issue in particular, they proposed the Attention Retractable transformer (ART) [25], where their method adds more pixel-level information as compared to the semantic-level information of its predecessors. They designed two self-attention blocks based on joint dense and sparse attention: dense attention blocks (DAB) that utilize fixed non-overlapping local windows, and sparse attention blocks (SAB) that use sparse grids to obtain tokens. Such changes allow ART to provide for longer distance residual connection between multiple Transformer encoders, enabling deep feature layers to reserve higher low-frequency information from shallow layers.

*Table I: A detailed comparison of State-of-the-art Transformer-based SR models.*

| Method | CONF | PSNR | SSIM | params (M) | FLOPs (G) | Base Network | Loss Fun | Paradigms | Training Datasets | Pre-trained |
|---|---|---|---|---|---|---|---|---|---|---|
| TTSR[18] | CVPR 20 | 25.87 | 0.784 | 6.42 | 185 | Transformer | $L_1 + \mathcal{L}_{per} + \mathcal{L}_{adv}$ | RefSR | CUFED5 | |

| | | | | | | | | | |
|---|---|---|---|---|---|---|---|---|---|
| IPT[19] | CVPR 21 | 27.26 | na | 114 | 33 | ViT | $L_1 + L_{con}$ | SISR | ImageNet | ✓ |
| SwinIR[20] | ICCV 22 | 27.45 | 0.825 | 11.90 | 215 | Swin | $L_1$ | SISR | DIV2k + | |
| DATSR[21] | ECCV 22 | 26.52 | 0.798 | 18 | na | UNet | $L_1 + L_{per} + L_{adv}$ | RefSR | CUFED5 | ✓ |
| ESRT[22] | CVPR 22 | 26.39 | 0.796 | 0.68 | 67.7 | Transformer | $L_1$ | SISR | DIV2k | |
| HAT[23] | CVPR 23 | 28.6 | 0.849 | 9.62 | 42.18 | ViT | $L_{pixel}$ | SISR | DF2k | ✓ |
| EDT[24] | IJCAI 23 | 27.46 | 0.824 | 11.6 | 37.6 | encoder-decoder | na | SISR | DF2k | ✓ |
| ART[25] | ICLR 23 | 27.77 | 0.832 | 16.55 | 300 | Transformer | $L_1$ | SISR | DF2K | |
| DAT[26] | ICCV 23 | 27.87 | 0.834 | 14.8 | 275.7 | Swin | $L_1$ | SISR | DF2k | ✓ |
| MAFT[27] | na | 20.08 | 0.837 | 14.07 | 258.9 | Transformer | $L_1$ | SISR | DF2k | |
| DRCT[28] | CVPR24 | 28.06 | 0.837 | 10.443 | 7.92 | Swin | $L_1 + L_2$ | SISR | DF2K | ✓ |

## 4. Conclusion & Future Directions

By identifying promising yet unexplored areas, the study lays the groundwork for future exploration and optimization of SR techniques. Despite recent developments, multiple challenges remain, such as high memory and computational demands, huge data dependency on Transformers, and low generalization across unseen degradation. Also, maintaining fine-grained textures and high-frequency details is still a challenge in complex scenes, as well as real-time SR generation, due to its high inference time, also requires further progress.

Therefore, there is a dire need for more efficient, lightweight, and adaptable SR models, which will decrease the inference time, and continued exploration of including classical methods like wavelets and interpolations with traditional ones like CNNs and GANs should be able to advance the field of SR. Also, models capable of handling diverse degradation types will ensure robustness for real-world use. By addressing these challenges and exploring emerging trends, transformer-based SR approaches will play a tremendous role in the advancement of the domain.